\documentclass[runningheads]{llncs}

\usepackage{graphicx}
\usepackage{amsmath,amssymb} 
\usepackage{color}
\usepackage[width=122mm,left=12mm,paperwidth=146mm,height=193mm,top=12mm,paperheight=217mm]{geometry}
\usepackage[compatibility=false]{caption}
\usepackage{subcaption}
\usepackage{makecell}
\usepackage{multirow}
\usepackage[export]{adjustbox}
\usepackage{mwe}

\begin{document}

\title{Improving Generalization via \\  Scalable Neighborhood Component Analysis}

\titlerunning{Scalable Neighborhood Component Analysis}

\author{Zhirong Wu\inst{1,2} \and
Alexei A. Efros\inst{1} \and
Stella X. Yu\inst{1}}

\authorrunning{Wu, Efros, Yu}

\institute{UC Berkeley / ICSI \and
Microsoft Research Asia
}

\newcommand{\norm}[1]{\left\lVert#1\right\rVert}

\maketitle

\begin{abstract}\let\thefootnote\relax\footnotetext{code \& models available: https://github.com/zhirongw/snca.pytorch}

Current major approaches to visual recognition 
follow an end-to-end formulation that classifies an input image into one of the pre-determined set of semantic categories.  Parametric softmax classifiers are a common choice for such a closed world with fixed categories, especially when big labeled data is available during training.
However, this becomes problematic for open-set scenarios where new categories are encountered with very few examples for learning a generalizable parametric classifier.
%
We adopt a non-parametric approach for visual recognition by optimizing feature embeddings instead of parametric classifiers. 
We use a deep neural network to learn the visual feature that preserves the neighborhood structure in the semantic space, based on the Neighborhood Component Analysis (NCA) criterion.
Limited by its computational bottlenecks,
we devise a mechanism to use augmented memory to scale NCA for 
large datasets and very deep networks.
Our experiments  deliver not only remarkable performance on ImageNet classification
for such a simple non-parametric method, 
but most importantly a more generalizable feature representation for sub-category discovery and few-shot recognition.

\end{abstract}

\section{Introduction}
\label{sec:intro}
 
Deep learning with end-to-end problem formulations has
reshaped visual recognition methods over the past few years.
The core problems of high-level vision, e.g.
recognition, detection and segmentation, are commonly 
formulated as classification tasks.
Classifiers are applied image-wise for recognition~\cite{krizhevsky2012imagenet},
region-wise for detection~\cite{ren2015faster}, and pixel-wise for segmentation~\cite{long2015fully}.
Classification in deep neural network is usually implemented as multi-way parametric softmax and assumes that the categories are fixed between learning and evaluation. 

However, such a ``closed-world" assumption does not hold for the open world,
where new categories could appear,
often with very few training examples.
For example, for face recognition~\cite{turk1991face,taigman2014deepface},
new identities should be recognized after just one-time occurrence.
Due to the open-set nature, one may want to generalize the feature embedding
instead of learning another parametric classifier.
A common practice for embedding 
is to simply chop off the softmax classification layer
from a pretrained network and take the last layer features.
However, such a transfer learning scheme is not optimal because
these features only make sense for a linear classification boundary in the training space,
most likely not for the new testing space. 
Instead of learning parametric classifiers,
we can learn an embedding to 
directly optimize a feature representation
which preserves distance metrics in a non-parametric fashion. 
Numerous works have investigated various loss functions
(e.g. contrastive loss~\cite{hadsell2006dimensionality}, triplet loss~\cite{hoffer2015deep,mensink2013distance}) and data sampling strategies~\cite{wu2017sampling} for improving the embedding performance.

Non-parametric embedding approaches
have also been applied to computer vision tasks other than face recognition.
Exemplar-based models have shown to be effective for
learning object classes~\cite{chum2007exemplar} and object detection~\cite{malisiewicz2011ensemble}.
These non-parametric approaches build associations between data instances~\cite{malisiewicz2008recognition},
and turn out to be useful for meta-knowledge transfer~\cite{malisiewicz2011ensemble} which would not be readily possible for parametric models.
So far, none of these non-parametric methods have become competitive in the state-of-the-art
image recognition benchmarks such as ImageNet classification~\cite{russakovsky2015imagenet} and MSCOCO object detection~\cite{lin2014microsoft}.
However, we argue that time might be right to revisit non-parametric methods to see if they could provide the generalization capabilities lacking in current approaches. 

 
We investigate a neighborhood approach for image classification
 by learning a feature embedding through deep neural networks.
 The core of our approach is a metric learning model based on
 Neighborhood Component Analysis (NCA)~\cite{goldberger2005neighbourhood}.
 For each training image, NCA computes its distance to all the other images in the embedding space.
 The distances can then be used to define a classification distribution according to the class labels.
 Batch training with all the images is computationally expensive,
 thereby making the original NCA algorithm difficult to scale to large datasets.
 Inspired by prior works~\cite{wu2018unsupervised,xiao2017joint}, we propose to store the embedding of images
 in the entire dataset in an augmented non-parametric memory.
 The non-parametric memory is not learned by stochastic gradient descent,
 but simply updated after each training image is visited.
 During testing, we build a $k$-nearest-neighbor (kNN) classifier based on the learned metrics.
 
 Our work makes three main contributions.
 1)  We scale up NCA
 to handle large-scale datasets and deep neural networks
 by using an augmented memory to store non-parametric embeddings.
 2)  We demonstrate that a nearest neighbor classifier
 can achieve remarkable performance
 on the challenging ImageNet classification benchmark, nearly on par with parametric methods.
 3)  Our learned feature, trained with the same embedding method,
 delivers improved generalization ability
 for new categories, which is desirable for sub-category discovery and few-shot recognition.

\section{Related Works}
\label{sec:related}

\textbf{Object Recognition.}
Object recognition is one of the holy grail problems in computer vision.
Most prior works cast recognition either as a category naming problem~\cite{deng2009imagenet,everingham2010pascal} or as a data association problem~\cite{malisiewicz2008recognition}.
Category naming assumes that all instances belonging to the same category are similar and that category membership is binary (either all-in, or all-out).  
Most of the research in this area is focused on designing better invariant category representations (e.g. bag-of-words~\cite{weber2000unsupervised}, pictorial models~\cite{felzenszwalb2005pictorial}).
On the other hand, data association approaches~\cite{chum2007exemplar,zhang2006svm,malisiewicz2008recognition,malisiewicz2009beyond} regard categories as data-driven entities emergent from connections between individual instances. 
Such non-parametric paradigms are informative and powerful for transferring knowledge
which may not be explicitly present in the labels.
In the era of deep learning, 
however, the performance of exemplar-based approaches hardly reaches the state-of-the-art for standard benchmarks on classification.
Our work revisits the direction of data association models,  learning an embedding representation
that is tailored for nearest neighbor classifiers.


\textbf{Learning with Augmented Memory.}
Since the formulation of LSTM~\cite{hochreiter1997long}, 
the idea of using memory for neural networks has been widely adopted for various tasks~\cite{hinton2012deep}.
Recent approaches on augmented memory fall into two camps.
One camp incorporates memory into neural networks as an end-to-end differentiable module~\cite{graves2014neural,memory_networks},
with automatic attention mechanism~\cite{santoro2016meta,vinyals2015pointer} for reading and writing.
These models are usually applied in knowledge-based reasoning~\cite{graves2014neural,vinyals2015pointer} and sequential prediction tasks~\cite{sukhbaatar2015end}.
The other camp treats memory as a non-parametric representation~\cite{vinyals2016matching,wu2018unsupervised,xiao2017joint}, where
the memory size grows with the data set size.
Matching networks~\cite{vinyals2016matching} explore few-shot recognition using augmented
memory, 
but their memory only holds the representations in current mini-batches of $5-25$ images.
Our memory is also non-parametric, in a similar manner as storing instances for unsupervised learning~\cite{wu2018unsupervised}.  The key distinction is that our approach learns the memory representation with millions of entries for supervised large-scale recognition.

\textbf{Metric Learning.}
There are many 
metric learning approaches~\cite{koestinger2012large,goldberger2005neighbourhood}, some achieving the state-of-the-art performance in image retrieval~\cite{wu2017sampling}, face recognition~\cite{schroff2015facenet,taigman2014deepface,wang2018cosface},
and person re-identification~\cite{xiao2017joint}.
In such problems, since the classes during testing are disjoint from those encountered during
 training, one can only make inference based on its feature representation, not on the subsequent linear classifier.
Metric learning learning encourages the minimization of intra-class variations
and the maximization inter-class variations, such as
contrastive loss~\cite{bromley1994signature,sohn2016improved}, triplet loss~\cite{hoffer2015deep}.
Recent works on few-shot learning~\cite{vinyals2016matching,snell2017prototypical}
also show the utility of metric learning, since it is difficult to optimize a parametric classifier
with very few examples.

\textbf{NCA.} Our work is built upon the original proposal of Neighborhood Component Analysis (NCA)~\cite{goldberger2005neighbourhood} and its
non-linear extension~\cite{salakhutdinov2007learning}.
In the original version~\cite{salakhutdinov2007learning}, the features
for the entire dataset needs to be computed at every step of the optimization,
making it computationally expensive and not scalable for large datasets.
Consequently, it has been mainly applied to small datasets such as MNIST or for dimensionality reduction~\cite{salakhutdinov2007learning}.
%
Our work is the first to demonstrate that NCA can be applied successfully to large-scale datasets.

\section{Approach}
\label{sec:approach}

We adopt a feature embedding framework for image recognition.
Given a query image $x$, we embed it into the feature space by $v  = f_\theta(x)$.
The function $f_\theta(\cdot)$ here is formulated as a deep neural network parameterized
by parameter $\theta$ learned from data $D$.
The embedding $v$ is then queried against a set of images in the search database $D'$, 
according to a similarity metric.
Images with the highest similarity scores are retrieved and information from
these retrieved images can be transferred to the image $x$.

Since the classification process does not rely on extra model parameters,
the non-parametric framework can naturally extend
to images in novel categories without any model fine-tuning.  Consider three settings of $D'$.
\begin{enumerate}
\setlength{\parskip}{0pt}\setlength{\leftskip}{-0.5em}\setlength{\itemsep}{0.5mm}
    \item 
When $D'=D$, i.e., the search database is the same as the training set, we have
closed-set recognition such as the ImageNet challenge.
\item 
When $D'$ is annotated with  
labels different from $D$,
we have open-set recognition such as sub-category discovery and few-shot recognition.
\item Even when $D'$ is completely unannotated, the metric can be useful for general content-based image retrieval.
\end{enumerate}

The key is how to learn such an embedding function $f_\theta(\cdot)$.
Our approach builds upon NCA~\cite{goldberger2005neighbourhood}
with some of our modifications.

\subsection{Neighborhood Component Analysis}

\subsubsection{Non-parametric formulation of classification.}
Suppose we are given a labeled dataset of $n$ examples $x_1, x_2, ... , x_n$ with
corresponding labels $y_1, y_2, ... , y_n$.  Each example $x_i$ is embedded into a feature vector $v_i = f_\theta(x_i)$.
We first define similarity $s_{ij}$ between instances $i$ and $j$ in the embedded space as cosine similarity. 
We further assume that the feature $v_i$ is $\ell_2$ normalized. Then,
\begin{equation} \label{eq:similarity}
s_{ij} = \text{cos} (\phi) = \frac{v^T_i }{\norm{v_i}\norm{v_j}} =  v^T_i v_j,
\end{equation}
where $\phi$ is the angle between vector $v_i$, $v_j$.
Each example $x_i$ selects example $x_j$ as its neighbor with probability $p_{ij}$ defined as,
\begin{equation} \label{eq:probability}
p_{ij} = \frac{\text{exp}(s_{ij} / \sigma )}{\sum_{k\neq i} \text{exp}(s_{ik} / \sigma)},  \quad p_{ii} = 0.
\end{equation}
Note that each example cannot select itself as neighbors, i.e. $p_{ii} = 0$.
The probability thus is called {\it leave-one-out} distribution on the training set.
Since the range of the cosine similarity is in $[-1, 1]$,
we add an extra parameter $\sigma$ to control the scale of the neighborhood.

Let $\Omega_i = \{j | y_j = y_i\}$ denote the indices of training images which share the
same label with example $x_i$.
Then the probability of example $x_i$ being correctly classified is,
\begin{equation} \label{eq:pi}
p_{i} = \sum_{j\in \Omega_i} p_{ij}.
\end{equation}
The overall objective is to minimize the expected negative log likelihood over the dataset,
\begin{equation} \label{eq:objective}
J =  \frac{1}{n} \sum_i J_i =  - \frac{1}{n} \sum_{i} \text{log} (p_i).
\end{equation}
Learning proceeds by directly optimizing the embedding without introducing additional model parameters.
It turns out that each training example depends on all the other exemplars in the dataset.
The gradients of the objective $J_i$  with respect to
$v_i$ is, 
\begin{equation} \label{eq:grad}
\frac{\partial J_i}{\partial v_i} = \frac{1}{\sigma} \sum_{k} p_{ik}v_k - \frac{1}{\sigma}\sum_{k\in \Omega_i} \tilde{p}_{ik} v_k,
\end{equation}
and $v_{j}$ where $j\neq i$ is, 
\begin{align} \label{eq:grad_j}
\frac{\partial J_i}{\partial v_j} = 
\begin{cases}
\frac{1}{\sigma}(p_{ij} - \tilde{p}_{ij} ) v_i, \quad &j\in \Omega_i \\
  \frac{1}{\sigma} p_{ij} v_i, \quad &j\notin \Omega_i 
\end{cases}
\end{align}
where $\tilde{p}_{ik} = p_{ik} / \sum_{j\in \Omega_i} p_{ij}$ is the normalized distribution within
the groundtruth category.

\subsubsection{Differences from parametric softmax.}
The traditional parametric softmax distribution is formulated as
\begin{equation} \label{eq:softmax}
p_c = \frac{\text{exp}(w^T_c v_i)}{\sum_j \text{exp} (w^T_j v_i)},
\end{equation}
where each category $c\in \left\{ 1,2,...,C \right\} $ has a parametrized prototype $w_c$ to represent itself. The maximum likelihood learning is to align all examples in the same category with the category prototype.
However, in the above NCA formulation, 
the optimal solution is reached when the probability $p_{ik}$ of negative examples ($k\notin \Omega_i$)
vanishes.
The learning signal does not enforce all the examples in the same category to align with the
current training example. 
The probability of some positive examples ($k \in \Omega_i$) can also
vanish so long as some other positives align well enough to $i$-{\it th} example.
In other words, the non-parametric formulation does not assume a single prototype for
each category, and such a flexibility allows learning to discover inherent structures 
when there are significant intra-class variations in the data.
Eqn~\ref{eq:grad} explains how each example contributes to the learning gradients.

\begin{figure}[t]
	\centering
	\includegraphics[width=1.0\linewidth]{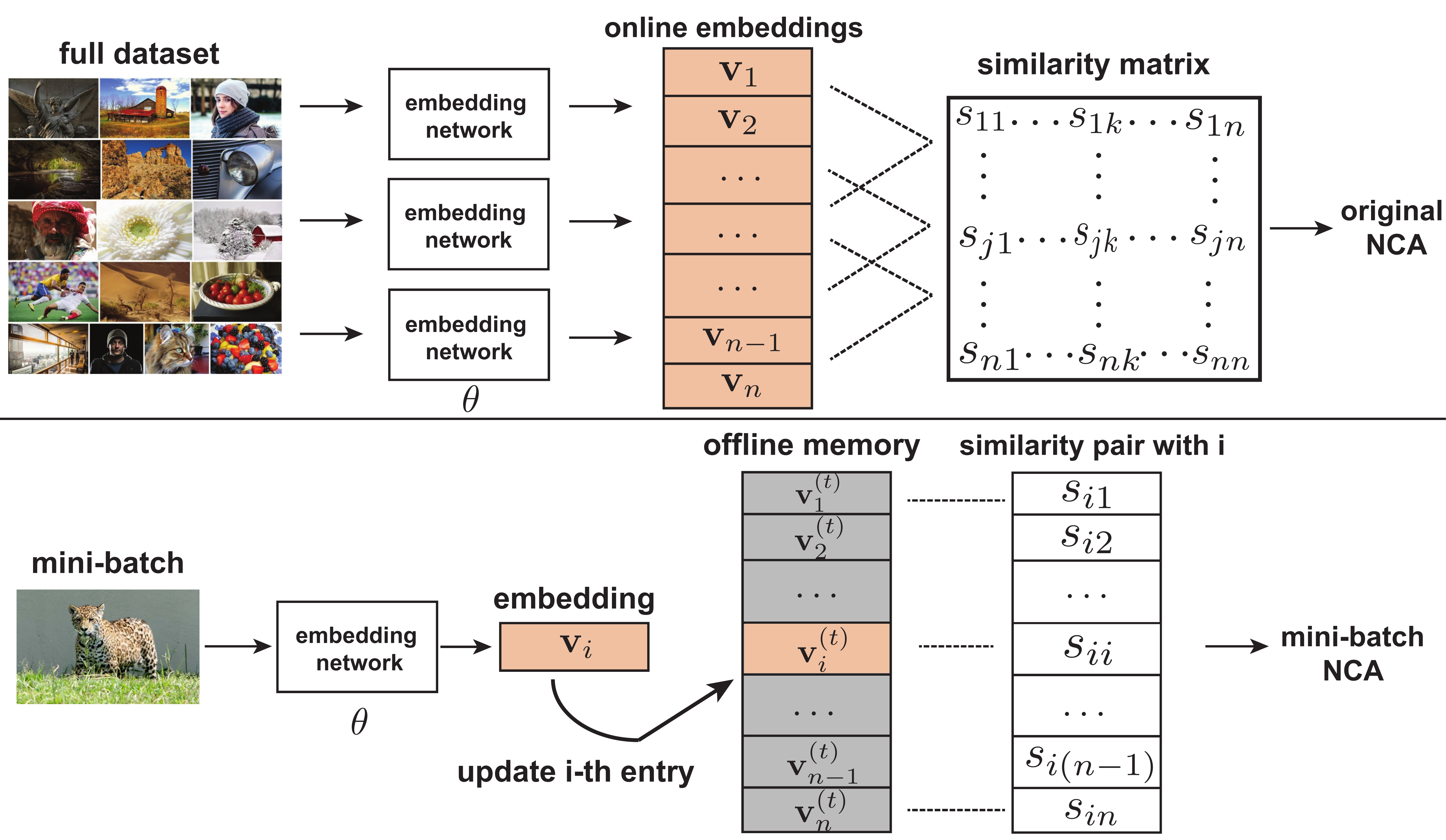}
	\caption{The original NCA needs to compute the feature embeddings 
		for the entire dataset for each optimization step.
		This is not scalable for large datasets and deep neural networks optimized
		with stochastic gradient descent.
		We overcome this issue by using an augmented memory to store offline embeddings
		forwarded from previous optimization steps.
		The online embedding is learned by back-propagation,
		while the offline memory is not.
	}
	\label{fig:nca}
\end{figure}

\subsubsection{Computational challenges for learning.}
Learning NCA even for a single objective term $J_i$ 
would require obtaining the embedding as well as 
gradients (Eqn~\ref{eq:grad} and Eqn~\ref{eq:grad_j})
in the entire dataset. 
This computational demand quickly becomes impossible to meet for large-scale dataset,
with a deep neural network learned via stochastic gradient descent.
Sampling-based methods such as triplet loss~\cite{taigman2014deepface}
can drastically reduce the computation by selecting a few neighbors.
However, hard-negative mining turns out to be crucial and
typical batch size with 1800 examples~\cite{taigman2014deepface} could still be impractical. 

We take an alternative approach to reduce the amount of computation.  We introduce two crude approximations.
\begin{enumerate}
\setlength{\parskip}{0pt}\setlength{\leftskip}{-0.5em}\setlength{\itemsep}{0.5mm}
\item 
We only perform gradient descent on $\partial J_i / \partial v_i$ as in Eqn~\ref{eq:grad},
but not on $\partial J_i / \partial v_{j}$, $ j\neq i$ as in Eqn~\ref{eq:grad_j}.
This simplification disentangles learning a single instance from learning among all the training instances,
making mini-batch stochastic gradient descent possible.
\item
Computing the gradient for $\partial J_i / \partial v_i$
 still requires the embedding of the entire dataset, which would be prohibitively expensive for each mini-batch update.
We introduce augmented memory 
to store the embeddings for approximation.  More details follow.

\end{enumerate}

\subsection{Learning with Augmented Memory}
We store the feature representation of the entire dataset as augmented non-parametric memory.
We learn our feature embedding network through stochastic gradient descent.
At the beginning of the $t$+1-{\it th} iteration, suppose the network parameter has the state $\theta^{(t)}$, and the non-parametric memory is in the form of
$M^{(t)} = \{v^{(t)}_1, v^{(t)}_2, ... , v^{(t)}_n\}$.
Suppose that the memory is roughly up-to-date with the parameter $\theta^{(t)}$ at iteration $t$.
This means the non-parametric memory is close to the features extracted from the data using parameter $\theta^{(t)}$, 
\begin{equation} \label{eq:memory}
v^{(t)}_i \approx f_{\theta^{(t)}}(x_i), \quad i=1,2,..., n.
\end{equation}
During the $t$+1-{\it th} optimization, for training instance $x_i$, we forward it through the embedding network $v_i = f_{\theta^{(t)}}(x_i)$, and calculate its gradient as in Eqn~\ref{eq:grad} but using the approximated embedding in the memory as,
\begin{equation} \label{eq:mem_grad}
\frac{\partial J_i}{\partial v_i} = \frac{1}{\sigma} \sum_{k} p_{ik}v^{(t)}_k - \frac{1}{\sigma}\sum_{k\in \Omega_i} \tilde{p}_{ik} v^{(t)}_k.
\end{equation}
Then the gradients of the parameter can be back-propagated,
\begin{equation} \label{eq:backprop}
\frac{\partial J_i}{\partial \theta} = \frac{\partial J_i}{\partial v_i} \cdot \frac{\partial v_i}{\partial \theta}.
\end{equation}
Since we have forwarded the $x_i$ to get the feature $v_i$, 
we update the memory for the training instance $x_i$ by the empirical weighted average~\cite{xiao2017joint},
\begin{equation} \label{eq:mem_update}
v^{(t+1)}_i \gets m \cdot v^{(t)}_i + (1-m) \cdot v_i.
\end{equation}

Finally, network parameter $\theta$ is updated and learned through 
stochastic gradient descent. If the learning rate is small enough, 
the memory can always be up-to-date
with the change of parameters. 
The non-parametric memory slot for each training image
is only updated once per learning epoch.
Though the embedding is approximately 
estimated, 
we have found it to work well in practice.

\subsection{Discussion on Complexity}
In our model, the non-parametric memory $M^{(t)}$, similarity metric $s_{ij}$, and probability density $p_{ij}$ may potentially require a large storage and pose computation bottlenecks.
We give an analysis of model complexity below.

Suppose our final embedding is of size $d=128$,  and we train our model on a typical large-scale dataset using $n=10^6$ images with a batch size of $b=256$.
Non-parametric memory $M$ requires $0.5$ GB $\left(O(dn)\right)$ of memory.
Similarity metric and probability density each requires $2$ GB $\left(O(bn)\right)$ of memory for storing the value and the gradient. In our current implementation, other intermediate variables used for computing the intra-class distribution require another $2$ GB $\left(O(bn)\right)$. In total, we would need $6.5$ GB for the NCA module.

In terms of time complexity, the summation in Eqn~\ref{eq:probability} and Eqn~\ref{eq:pi} across the whole dataset becomes the bottleneck in NCA. However, in practice with a GPU implementation, the NCA module
takes a reasonable $30\%$ amount of extra time with respect to the backbone network. 
During testing, exhaustive nearest neighbor search with one million entries is also reasonably fast.
The time it takes is negligible with respect to the forward passing through the backbone network.

The complexity of our model scales linearly with the  training size set.  Our current implementation can deal with datasets at the ImageNet scale, but cannot scale up to 10 times more data based on the above calculations.  A possible strategy to handle bigger data is to subsample a few neighbors instead of the entire training set. Sampling would help reduce the linear time complexity to a constant.
For nearest neighbor search at the run time, computation complexity can be mitigated with proper data structures such as
ball-trees~\cite{friedman1977algorithm} and quantization methods~\cite{jegou2011product}.

\section{Experiments}

We conduct experiments to investigate whether
our non-parametric feature embedding can
perform well in the closed-world setting, and more importantly whether it can improve generalization in the open-world setting.

First, we evaluate the learned metric on the large-scale ImageNet ILSVRC challenge~\cite{russakovsky2015imagenet}.
Our embedding achieves competitive recognition accuracy with {\it k}-nearest neighbor classifiers
using the same ResNet architecture.
Secondly, we study an important property of our
representation for sub-category discovery,
when the model trained with only coarse annotations
is transferred for fine-grained label prediction.
Lastly, we study how our learned metric can be 
transferred and applied to unseen object categories for few-shot recognition.

\subsection{Image Classification}
We study the effectiveness of our non-parametric representation for visual recognition on ImageNet ILSVRC dataset.
We use the parametric softmax classification networks as our baselines.

\noindent
\textbf{Network Configuration}.
We use the ConvNet architecture ResNet\cite{he2016deep} as the backbone for the feature embedding network.
We remove the last linear classification layer of the original ResNet and append another
linear layer which projects the feature to a low dimensional 128 space.
The 128 feature vector is then $\ell_2$ normalized and fed to NCA learning.
Our approach does not induce extra parameters for the embedding network.

\noindent
\textbf{Learning Details}.
During training, we use an initial learning rate of 0.1 and 
drops 10 times smaller every 40 epochs for a total of 130 epochs.
Our network converges a bit slower than the baseline network, in part due to the approximated updates for the non-parametric memory.
We set the momentum for updating the memory with $m=0.5$ at the start of learning, and gradually increase to $m=0.9$ at the end of learning.
We use a temperature parameter $\sigma = 0.05$ in the main results.
All the other optimization details and hyper-parameters remain the same with the baseline approach.
We refer the reader to the PyTorch implementation~\cite{paszkepytorch} of ResNet for details.
During testing, we use a weighted {\it k} nearest neighbor classifier for classification.
Our results are insensitive to parameter $k$; generally any $k$ in the range of $5 - 50$ gives very similar results.
We report the accuracy with $k=1$ and $k=30$ using single center crops.

\begin{table}[t]
	\setlength{\tabcolsep}{2.2pt}
	\centering
	\caption{Top-1 classification rate on ImageNet validation set using {\it k}-nearest neighbor classifiers. 
	}
	\footnotesize
	\begin{tabular}{c|c|c|c}
		\Xhline{2\arrayrulewidth} 
		\multicolumn{4}{c}{ResNet18} \\
		\Xhline{2\arrayrulewidth} 
		Feature & $d$ & $k$=1 & $k$=30 \\
		\hline
		Baseline & 512 & 62.91 & 68.41 \\
		\hline
		+PCA & 128 & 60.43 & 66.26  \\
		\Xhline{2\arrayrulewidth} 
		Ours & 128 & 67.39 & \textbf{70.58} \\
		\Xhline{2\arrayrulewidth}
	\end{tabular}
	\begin{tabular}{c|c|c|c}
		\Xhline{2\arrayrulewidth} 
		\multicolumn{4}{c}{ResNet34} \\
		\Xhline{2\arrayrulewidth} 
		Feature & $d$ & $k$=1 & $k$=30 \\
		\hline
		Baseline & 512 & 67.73 & 72.32 \\
		\hline
		+PCA & 128 & 65.58 & 70.67  \\
		\Xhline{2\arrayrulewidth} 
		Ours & 128 & 71.81 & \textbf{74.43} \\
		\Xhline{2\arrayrulewidth}
	\end{tabular}
	\begin{tabular}{c|c|c|c}
		\Xhline{2\arrayrulewidth} 
		\multicolumn{4}{c}{ResNet50} \\
		\Xhline{2\arrayrulewidth} 
		Feature & d & $k$=1 & $k$=30 \\
		\hline
		Baseline & 2048 & 71.35 & 75.09 \\
		\hline
		+PCA & 128 & 69.72 & 73.69  \\
		\Xhline{2\arrayrulewidth} 
		Ours & 128 & 74.34 & \textbf{76.67} \\
		\Xhline{2\arrayrulewidth}
	\end{tabular}
	\label{table:imagenet}
\end{table}

\begin{table}[t]
\begin{minipage}{0.5\linewidth}
	\setlength{\tabcolsep}{3pt}
	\centering
	\caption{Performance comparison of our method with parametric softmax.
	}
	\footnotesize
	\begin{tabular}{c|c|c|c|c}
		\Xhline{2\arrayrulewidth} 
		Feature & \multicolumn{2}{c|}{baseline} & \multicolumn{2}{c}{ours} \\
		\hline
		 & top-1 & top-5 & top-1 & top-5 \\
		\hline
		ResNet18 & 69.64 & 88.98 & \textbf{70.58} & \textbf{89.38}\\
		\hline
		ResNet34 & 73.27 & \textbf{91.43} & \textbf{74.43} & 91.35\\
		\hline
		ResNet50 & 76.01 & \textbf{92.93} & \textbf{76.67} & 92.84\\
		\Xhline{2\arrayrulewidth}
	\end{tabular}
	\label{table:comparison}
\end{minipage}
\begin{minipage}{0.5\linewidth}
    \setlength{\tabcolsep}{3pt}
	\centering
	\caption{Ablation study on the feature size and the temperature parameter.}
	\footnotesize
	\begin{tabular}{c|c|c}
		\Xhline{2\arrayrulewidth} 
		$d$ & $k$=1 & $k$=30 \\
		\hline
		256 & 67.54 & 70.71 \\
		\hline
		128 & 67.39 & 70.59  \\
		\hline 
		64 & 65.32 &  69.54 \\
		\hline
		32 & 64.83 &  68.01 \\
		\Xhline{2\arrayrulewidth}
	\end{tabular}
	\quad
	\begin{tabular}{c|c|c}
		\Xhline{2\arrayrulewidth} 
		$\sigma$ & $k$=1 & $k$=30 \\
		\hline
		0.1 & 63.87 & 67.93 \\
		\hline
		0.05 & 67.39 & 70.59  \\
		\hline 
		0.03 & 66.98 & 70.33 \\
		\hline
		0.02 & N/A & N/A \\
		\Xhline{2\arrayrulewidth}
	\end{tabular}
	\label{table:ablation}
\end{minipage}
\end{table}

\noindent
\textbf{Main Results}.
Table~\ref{table:imagenet} and Table~\ref{table:comparison} summarize our results in comparison with the features learned by parametric softmax.
For baseline networks,
we extract the last layer feature and evaluate it with the same {\it k} nearest neighbor classifiers.
The similarity between features is measured by cosine similarity.
Classification evaluated with nearest neighbors leads to 
a decrease of $6\%-7\%$ accuracy with $k=1$, and $1\%-2\%$ accuracy with $k=30$.
We also project the baseline feature to $128$ dimension with PCA for evaluation.
This reduction leads to a further $2\%$ decrease in performance, suggesting that the features learned by parametric classifiers do not 
work equally well with nearest neighbor classifiers.
With our model, we achieve a $3\%$ improvement over the baseline using $k=1$.
At $k=30$, we have even slightly better results than the parametric classifier:
Ours are $1.1\%$ higher on ResNet34, and $0.7\%$ higher on ResNet50.
We also find that predictions from our model disagree with 
the baseline on $15\%$ of the validation set,
indicating a significantly different representation has been learned.

\begin{figure}[t]
	\centering
	\includegraphics[width=1.0\linewidth]{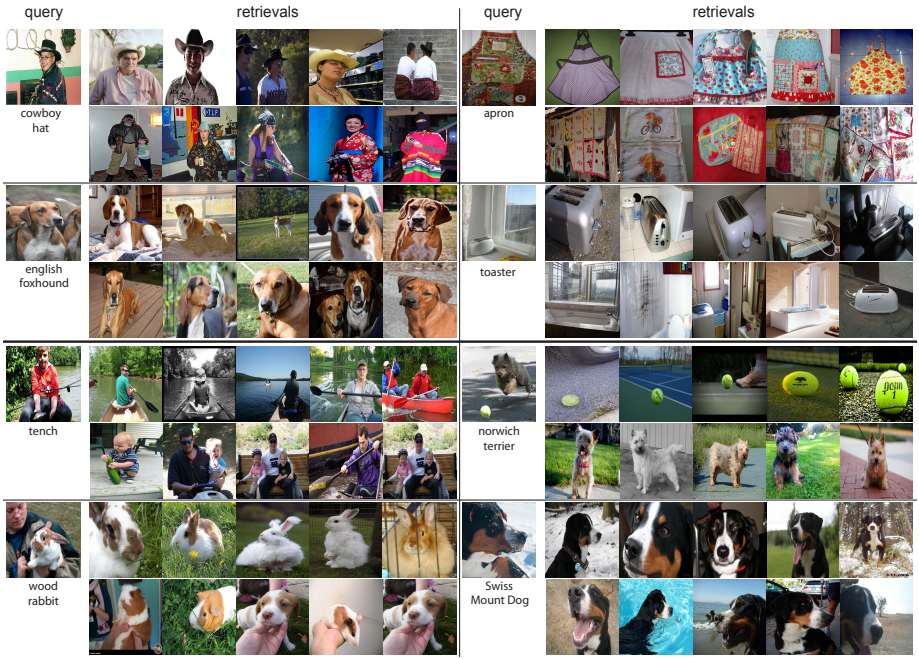}
	\caption{Given a query, the figure shows 5 nearest neighbors from our model (1st row) and from the baseline model (2nd row).
		Top four examples show the successful cases and bottom four show the failure cases.
	}
	\label{fig:comparisons}
\end{figure}

Figure~\ref{fig:comparisons} shows nearest neighbor retrieval comparisons. 
The upper four examples are our successful retrievals and the lower four are failure retrievals.
For the failure cases, our model has trouble either when
there are multiple objects in the same scene, or when  the task becomes too difficult with fine-grained categorization.
For the four failure cases, our model predictions are
``paddle boat", ``tennis ball", ``angora rabbit", ``appenzeller" respectively.

\noindent
\textbf{Ablation study on model parameters}.
We investigate the effect of the feature size and the temperature parameter in Table~\ref{table:ablation}.
For the feature size, 128 features and 256 features produce very similar results.
We start to see performance degradation as the size is dropped lower than 64.
For the temperature parameter, a lower temperature which induces smaller neighborhoods generally produces  better results. 
However, the network does not converge if the temperature is too low, e.g., $\sigma = 0.02$.

\begin{table}[t]
	\setlength{\tabcolsep}{3pt}
	\centering
	\caption{Top-1 induction accuracy on CIFAR100 and ImageNet1000 using model pretrained on
		CIFAR20 and ImageNet127. Numbers are reported with {\it k} nearest neighbor classifiers.}
	\footnotesize
	\begin{tabular}{c|c|c}
		\Xhline{2\arrayrulewidth} 
		\multicolumn{3}{c}{CIFAR} \\
		\Xhline{2\arrayrulewidth} 
		Task & 20 classes & 100 classes \\
		\hline
		Baseline & 81.53 & 54.17\\
		\Xhline{2\arrayrulewidth} 
		Ours  & 81.42 & \textbf{62.32}\\
		\Xhline{2\arrayrulewidth}
	\end{tabular}
	\begin{tabular}{c|c|c}
		\Xhline{2\arrayrulewidth} 
		\multicolumn{3}{c}{ImageNet} \\
		\Xhline{2\arrayrulewidth} 
		Task &  127 classes & 1000 classes \\
		\hline
		Baseline & 81.48  & 48.07\\
		\hline
		Ours & 81.62 & \textbf{52.75} \\
		\Xhline{2\arrayrulewidth}
	\end{tabular}
	
	\label{table:induction}
\end{table}

\begin{figure}[t]
	\centering
	\includegraphics[height=90pt]{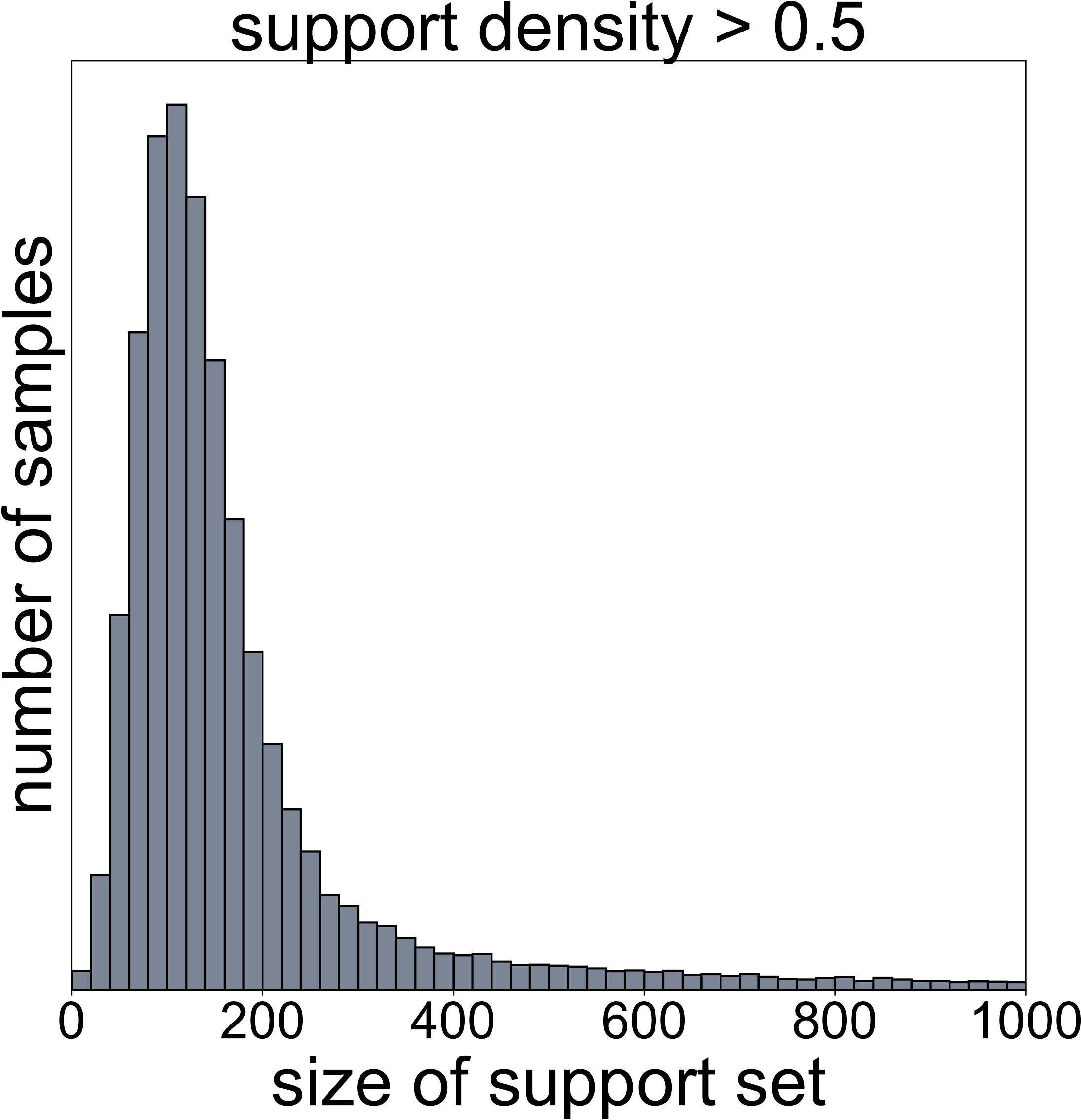}
	~
	\includegraphics[height=90pt]{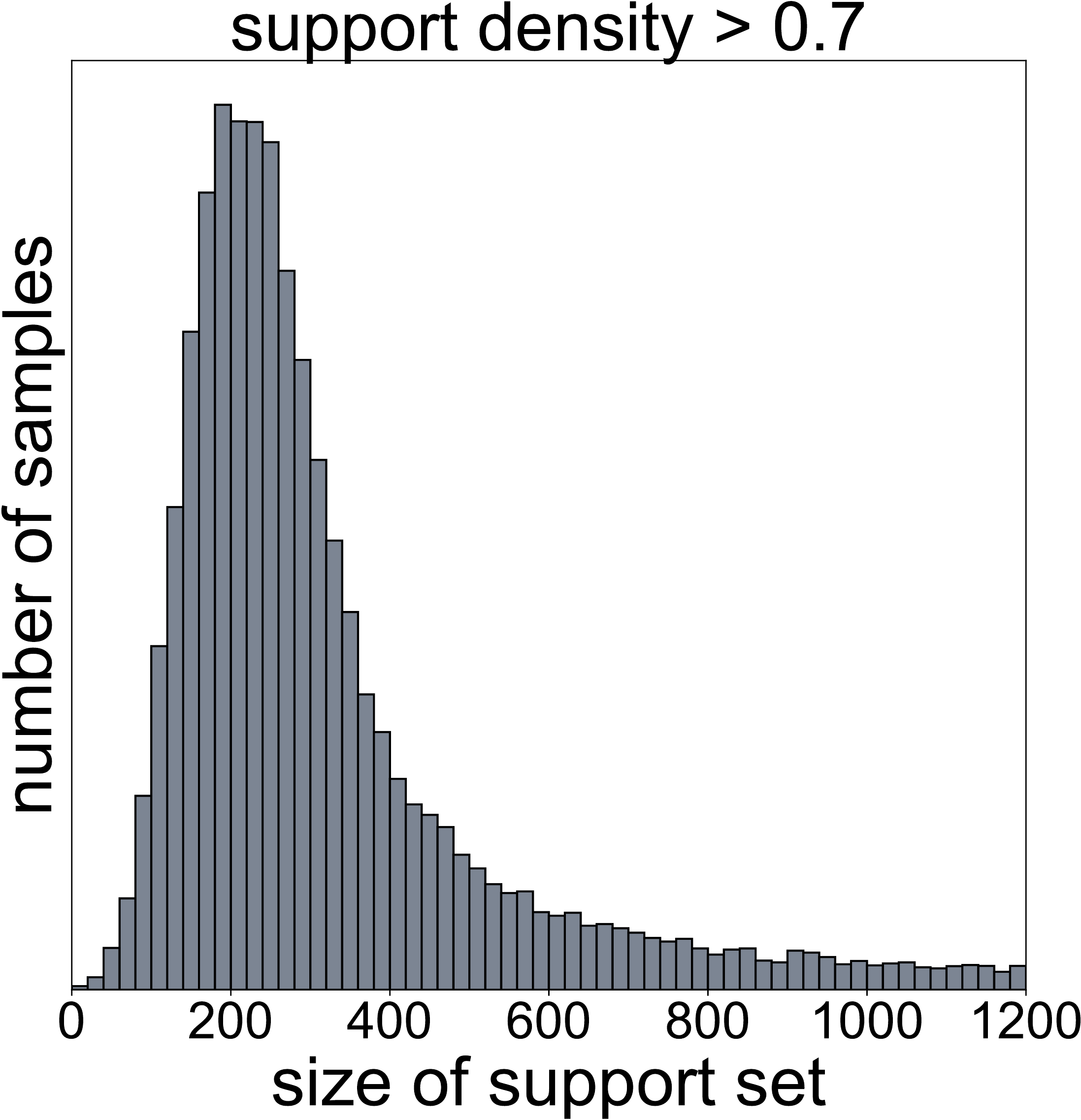}
	~
	\includegraphics[height=90pt]{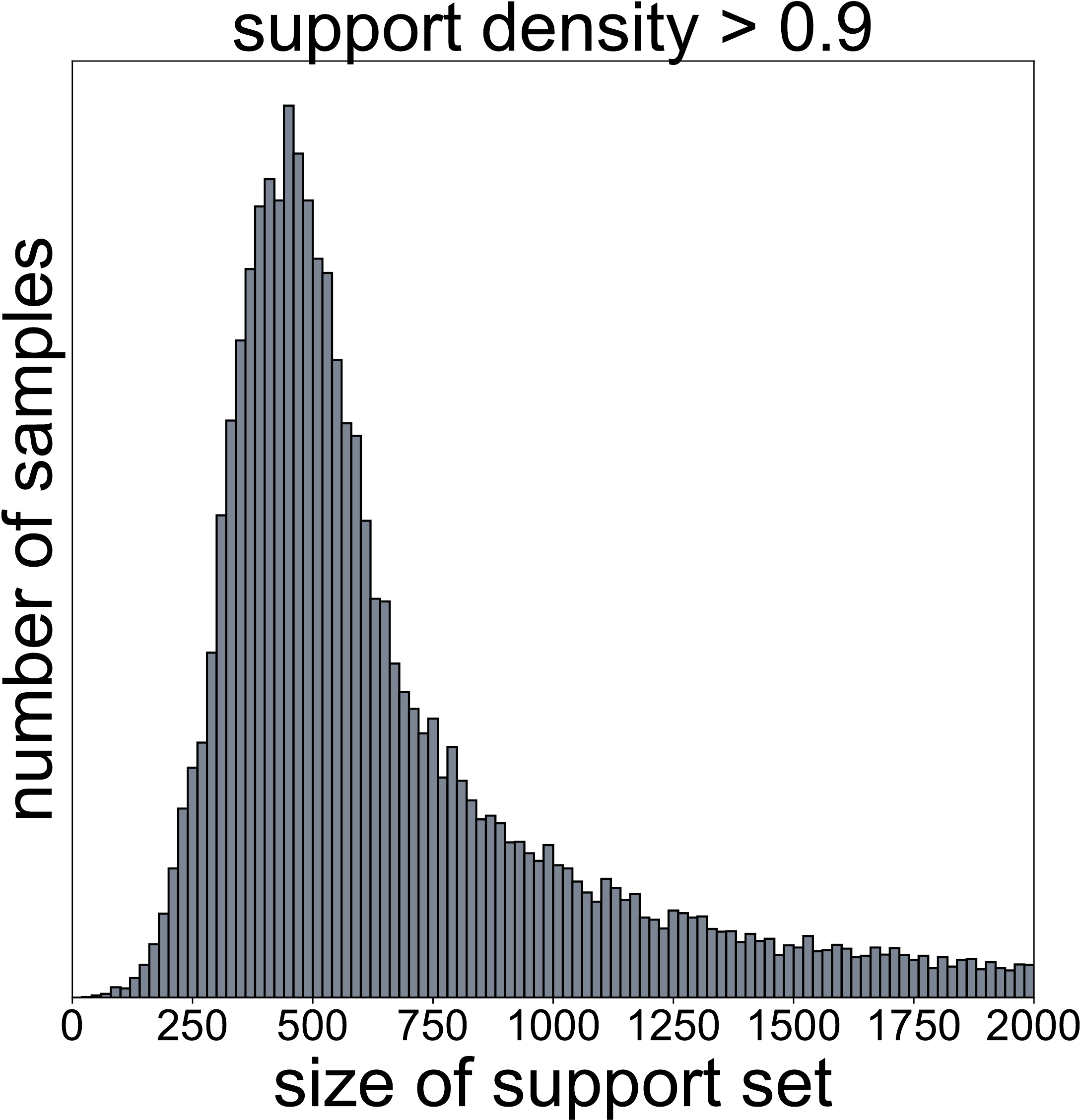}
	\caption{Histogram of the size of support set in the ImageNet validation set given
		various support density thresholds.
	}
	\label{fig:support_set}
\end{figure}

\subsection{Discovering Sub-Categories}

\begin{figure}[t]
	\centering
	\includegraphics[width=1.0\linewidth]{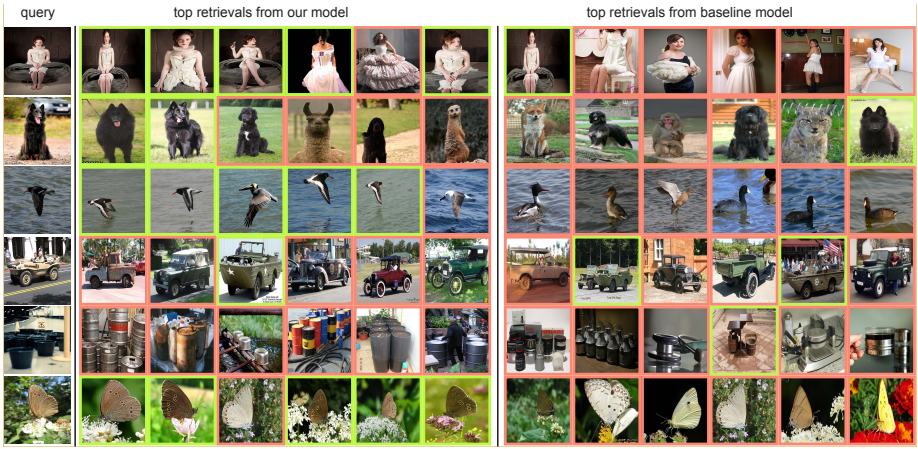}
	\caption{
		Nearest neighbors from the models trained with ImageNet 127 classes
		and evaluated on the fine-grained 1000 classes.
		Correct retrievals are boxed with green outlines and wrong retrievals are with orange.
	}
	\label{fig:coarse}
\end{figure}
\begin{figure}[t]
	\centering
	\raisebox{-0.5\height}{\includegraphics[width=0.99\linewidth]{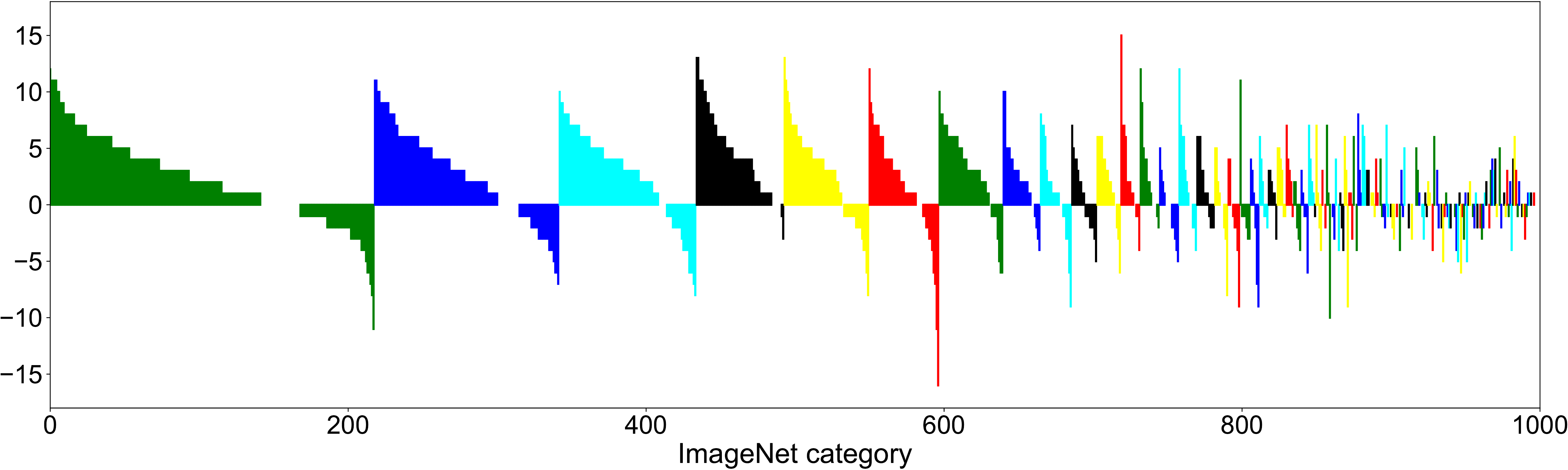}}
	\caption{
		Results for sub-category discovery on ImageNet.
		$x$ axis scans through the fine-grained 1000 ImageNet categories.
		Each recycled color represents a coarse category.
		All coarse categories are sorted with decreasing order
		in terms of the number of sub-categories.
		$y$ axis indicates the prediction gains of
		our model against the baseline model.
		Within each coarse category, the prediction gains for sub-categories are also sorted in a decreasing order.
	}
	\label{fig:coarse_compare}
\end{figure}

Our non-parametric formulation of classification does not assume a single prototype for each category.
Each training image $i$ only has to look for a few supporting neighbors~\cite{scholkopf2001estimating} to embed the features.
We refer nearest neighbors whose probability density $\sum_{j} p_{ij}$
sum over a given threshold as a support set for $i$.
In Figure~\ref{fig:support_set}, we plot 
the histograms over the size of the support set for support density thresholds  $0.5$, $0.7$ and $0.9$. 
We can see most of the images only
depend on around $100-500$ neighbors, which are a lot less than 1,000 images per category in ImageNet.
These statistics suggest that our learned representation allows sub-categories to develop automatically.

The ability to discover sub-categories is of great importance for feature learning, as there are always intra-class
variations no matter how we define categories.
For example, even for the finest level of object species, we can further
define object pose as sub-categories.

To quantitatively measure the performance of sub-category discovery, 
we consider the experiment of learning the feature embedding using coarse-grained object labels,
and evaluating the embedding using fine-grained object labels.
We can then measure how well feature learning discovers variations within categories.
We refer this classification performance as induction accuracy as in~\cite{huh2016makes}.
We train the network with the baseline parametric softmax
and with our non-parametric NCA using the same network architecture.
To be fair with the baseline, we evaluate the feature from the penultimate layer
from both networks. 
We conduct the experiments on CIFAR and ImageNet, and their results are summarized in 
Table~\ref{table:induction}.

\noindent
\textbf{CIFAR Results}.
CIFAR100 ~\cite{krizhevsky2009learning} images have both fine-grained annotations in 100 categories and coarse-grained annotations in 20 categories.
It is a proper testing scenario for evaluating sub-category discovery.
We study sub-category discovery by transferring representations learned from 20 categories
to 100 categories.
The two approaches exhibit similar
classification performances on the 20 category setting.
However, when transferred to CIFAR100 using {\it k} nearest neighbors, 
baseline features suffer a big loss, with $54.17\%$ top-1 accuracy on 100 classes.
Fitting a linear classifier for the baseline features gives an improved $58.66\%$ top-1 accuracy.
Using {\it k} nearest neighbor classifiers, our features
are $8\%$ better than the baselines, achieving a $62.32\%$ recognition accuracy.

\noindent
\textbf{ImageNet Results}.
As in~\cite{huh2016makes}, 
we use 127 coarse categories by clustering the 1000 categories in a top-down fashion by fixing
the distances of the nodes from the root node in the WordNet tree.
There are 65 of the 127 classes present in the original 1000 classes.
The other 62 classes are parental nodes in the ImageNet hierarchical word tree. 
The two models achieve similar classification performance  ($81\%-82\%$) on the original 127 categories. 
When evaluated with 1000 class annotations, our representation
is about $5\%$ better than the baseline features. 
The baseline performance can be improved to $52.0\%$ 
by fitting another linear classifier on the 1000 classes.

\noindent
\textbf{Discussions}.
Our approach is able to preserve visual structures
which are not explicitly presented in the supervisory signal.
In Figure~\ref{fig:coarse},
we show nearest neighbor examples compared with the baseline features.
For all the examples shown here,
the ground-truth fine-grained category does not exist in the training categories.
Thus the model has to discover sub-categories in order to recognize the objects.
We can see our representation preserves apparent visual similarity (such as color and pose information) better,
and is able to associate the query with correct exemplars for accurate recognition.
For example, our model finds similar birds hovering above water in the third row,
and finds butterflies of the same color in the last row.
In Figure~\ref{fig:coarse_compare} we further show the prediction gains
for each class.
Our model is particularly stronger for main sub-categories with rich intra-class variations.

\begin{table}[t]
	\setlength{\tabcolsep}{3pt}
	\centering
	\caption{Few-shot recognition on Mini-ImageNet dataset.}
	\footnotesize
	\begin{tabular}{c|c|c|c|c|c|c}
		\Xhline{2\arrayrulewidth} 
		\multirow{2}{*}{Method} & \multirow{2}{*}{Network} &  \multirow{2}{*}{FineTune} & \multicolumn{2}{c|}{5-way Setting} & \multicolumn{2}{c}{20-way Setting} \\
		\cline{4-7}
		&  &  & 1-shot & 5-shot & 1-shot & 5-shot \\
		\hline
		NN Baseline~\cite{vinyals2016matching} & Small & No & 41.1$\pm$0.7 & 51.0$\pm$0.7 & - & - \\
		\hline
		Meta-LSTM~\cite{ravi2016optimization} & Small  & No & 43.4$\pm$0.8 & 60.1$\pm$0.7 & 16.7$\pm$0.2 & 26.1$\pm$0.3\\
		\hline
		MAML~\cite{finn2017model} & Small & Yes & 48.7$\pm$0.7 & 63.2$\pm$0.9 & 16.5$\pm$0.6 & 19.3$\pm$0.3 \\
		\hline
		Meta-SGD~\cite{li2017meta} & Small & No & 50.5$\pm$1.9 & 64.0$\pm$0.9 & 17.6$\pm$0.6 & 28.9$\pm$0.4\\
		\hline
		Matching Net~\cite{vinyals2016matching} & Small & Yes & 46.6$\pm$0.8  & 60.0$\pm$0.7  & - & - \\
		\hline
		Prototypical~\cite{snell2017prototypical} & Small & No & 49.4$\pm$0.8 & \textbf{68.2}$\pm$0.7 & - & - \\
		\hline
		RelationNet~\cite{sung2017learning} &  Small & No & \textbf{51.4}$\pm$0.8 & 61.1$\pm$0.7 & - & - \\
		\hline
		Ours & Small & No & 50.3$\pm$0.7 & 64.1$\pm$0.8 & \textbf{23.7}$\pm$0.4  & \textbf{36.0}$\pm$0.5\\
		\Xhline{2\arrayrulewidth}
		SNAIL~\cite{mishra2017meta} & Large & No  & 55.7$\pm$1.0 & 68.9$\pm$0.9 & - & - \\
		\hline
		RelationNet~\cite{sung2017learning} & Large & No & 57.0$\pm$0.9 & 71.1$\pm$0.7 & - & - \\
		\hline
		Ours & Large & No & \textbf{57.8}$\pm$0.8 & \textbf{72.8}$\pm$0.7 & \textbf{30.5}$\pm$0.5 & \textbf{44.8}$\pm$0.5\\
		\Xhline{2\arrayrulewidth}
	\end{tabular}
	\label{table:few-shot}
\end{table}

\subsection{Few-shot Recognition}

Our feature embedding method learns a meaningful metric among images.
Such a metric can be directly applied to new image categories which have not been seen during training.
We study the generalization ability of our method for few-shot object recognition.

\noindent
\textbf{Evaluation Protocol}.
We use the 
mini-Imagenet dataset~\cite{vinyals2016matching}, which consists of 60,000 colour images and 100 classes (600 examples per class). 
We follow the split introduced previously~\cite{ravi2016optimization}, with 64, 16, and 20 classes for training, validation and testing.
We only use the validation set for tuning model parameters.
During testing, we create the testing episodes by randomly sampling a set of observation
and query pairs. The observation consists of $c$ classes ($c$-way)  and $s$ images ($s$-shot) per class.
The query is an image from one of the$c$ classes.
Each testing episode provides the task to predict the class of query image
given $c\times s$ few shot observations.
We create $3,000$ episodes for testing and report the average results.

\noindent
\textbf{Network Architecture}.
We conduct experiments on two network architectures.
One is a shallow network which receives small $84\times84$ input images. 
It has 4 convolutional blocks, each with a $3\times 3 \times 64$ convolutional layer,
a batch normalization layer, a ReLU layer, and a max pooling layer.
A final fully connected layer maps the feature for classification.
This architecture is widely used in
previous works~\cite{finn2017model,vinyals2016matching} for evaluating few-shot recognition.
The other is a deeper version with ResNet18 and larger $224\times 224$ image inputs.
Two previous works~\cite{mishra2017meta,sung2017learning}
have reported their performance with similar ResNet18 architectures.

\begin{figure}[t]
	\centering
	\includegraphics[width=1.0\linewidth]{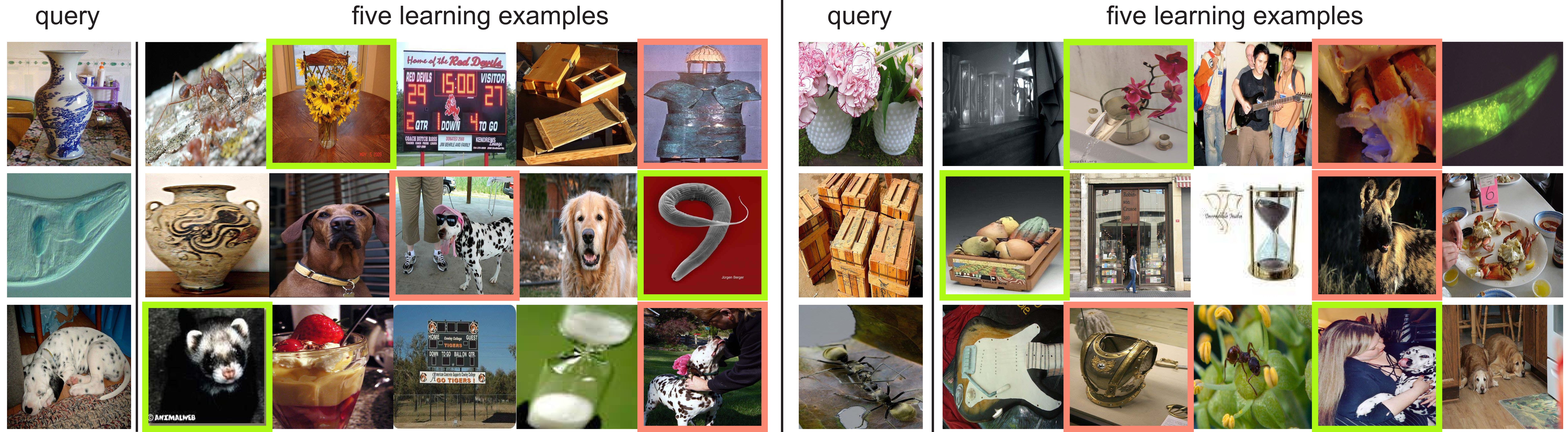}
	\caption{Few shot learning examples in mini-Imagenet test set.
		Given one shot for each five categories, the model
		predicts the category for the new query image.
		Our prediction is boxed with green and the baseline prediction is with orange.
	}
	\label{fig:fewshot}
\end{figure}

\noindent
\textbf{Results}.
We summarize our results in Table~\ref{table:few-shot}. 
We train our embedding on the training set, 
and apply the representation from the penultimate layer for evaluation.
Our current experiment does not fine-tune a local metric per episode,
though such adaptation would potentially bring additional improvement.
As with the previous experiments, we use {\it k} nearest neighbors for classification.
We use $k=1$ neighbor for the 1-shot scenario, and $k=5$ for the 5-shot scenario.

For the shallow network setting,
while our model is on par with the prototypical network~\cite{snell2017prototypical},
and RelationNet~\cite{sung2017learning}, our method is far more generic.

For the deeper network setting,
we achieve the state-of-the-art results for this task.
MAML~\cite{finn2017model} suggests going deeper 
does not necessarily bring better results for meta learning.
Our approach provides a counter-example:
Deeper network architectures can in fact bring  significant gains with proper metric learning.

Figure~\ref{fig:fewshot} shows visual examples of our
predictions compared with the baseline trained with softmax classifiers.

\section{Summary}
We present a non-parametric neighborhood approach for visual recognition.
We learn a CNN to embed images
into a low-dimensional feature space, where the distance metric between
images preserves the semantic structure of categorical labels according to the NCA criterion.
We address NCA's computation demand by learning with an
external augmented memory, thereby making NCA scalable for large datasets and deep
neural networks.
Our experiments  deliver not only remarkable performance on ImageNet classification
for such a simple non-parametric method, 
but most importantly a more generalizable feature representation for sub-category discovery and few-shot recognition.
In the future, it's worthwhile to re-investigate non-parametric methods for other
visual recognition problems such as detection and segmentation.

\section*{Acknowledgements}
This work was supported in part by Berkeley DeepDrive.
ZW would like to thank Yuanjun Xiong for helpful discussions.

\clearpage

\bibliographystyle{splncs04}
\bibliography{egbib}

\end{document}